%% file: main.tex
\documentclass[sigconf]{acmart}
\AtBeginDocument{%
  }

\copyrightyear{2026}
\acmYear{2026}
\setcopyright{cc}
\setcctype{by}
\acmConference[SIGMOD Companion '26] {Companion of the International Conference on Management of Data}{May 31--June 5, 2026}{Bengaluru, India.}
\acmBooktitle{Companion of the International Conference on Management of Data (SIGMOD Companion '26), May 31--June 5, 2026, Bengaluru, India}
\acmISBN{979-8-4007-2450-3/2026/05}
\acmDOI{10.1145/3788853.3801608}

\newcommand{\hide}[1]{}

\newcommand{\para}[1]{\vspace{1mm}\noindent\textbf{#1.}}

\input{macros}
\usepackage{listings}
\usepackage{xcolor}
\usepackage{seqsplit}
 \usepackage{balance} 
\begin{document}

\title{BDI-Kit Demo: A Toolkit for Programmable and Conversational Data Harmonization}


\author{Roque Lopez}
\affiliation{%
  \institution{New York University}
  \city{New York}
  \state{NY}
  \country{USA}}
\email{rlopez@nyu.edu}
\author{Yurong Liu}
\affiliation{%
  \institution{New York University}
  \city{New York}
  \state{NY}
  \country{USA}}
\email{yurong.liu@nyu.edu}
\author{Christos Koutras}
\affiliation{%
  \institution{New York University}
  \city{New York}
  \state{NY}
  \country{USA}}
\email{christos.koutras@nyu.edu}
\author{Juliana Freire}
\affiliation{%
  \institution{New York University}
  \city{New York}
  \state{NY}
  \country{USA}}
\email{juliana.freire@nyu.edu}



\input{sections/00_abstract}

\ccsdesc[500]{Information systems~Information integration}

\keywords{Data Harmonization, Schema Matching, Value Matching, AI Agents}


\maketitle
\input{sections/01_introduction}
\input{sections/02_system}
\input{sections/03_demonstration}
\input{sections/04_acknowledgments}

\bibliographystyle{ACM-Reference-Format}
\balance
\bibliography{references}

\end{document}

%% file: macros.tex
\def\withnotes{1} 

\definecolor{HighlightColor}{rgb}{0, 0, 255}
\newcommand{\revision}[1]{{\color{HighlightColor}#1}}
\renewcommand{\revision}[1]{#1}

\ifnum\withnotes=1    
    \newcommand{\roque}[1]{\textcolor{teal}{\textbf{Roque}: #1}}
    \newcommand{\yurong}[1]{\textcolor{orange}{\textbf{Yurong}: #1}}
\else
    \newcommand{\roque}[1]{}
    \newcommand{\yurong}[1]{}
\fi

%% file: sections/00_abstract.tex
\begin{abstract}
Data harmonization remains a major bottleneck for integrative analysis due to heterogeneity in schemas, value representations, and domain-specific conventions. BDI-Kit provides an extensible toolkit for schema and value matching. It exposes two complementary interfaces tailored to different user needs: a Python API enabling developers to construct harmonization pipelines programmatically, and an AI-assisted chat interface allowing domain experts to harmonize data through natural language dialogue. This demonstration showcases how users interact with BDI-Kit to iteratively explore, validate, and refine schema and value matches through a combination of automated matching, AI-assisted reasoning, and user-driven refinement. We present two scenarios: (i) using the Python API to programmatically compose primitives, examine intermediate outputs, and reuse transformations; and (ii) conversing with the AI assistant in natural language to access BDI-Kit's capabilities and iteratively refine outputs based on the assistant's suggestions.
\end{abstract}

%% file: sections/01_introduction.tex
\section{Introduction}

Integrating datasets is essential for large-scale analysis, yet remains challenging due to heterogeneity in schemas and value formats~\citep{magneto@vldb2025}. Despite decades of research in schema and value matching~\citep{doan2012integrationBook,cafarellaIntegrationonweb2009,miller2018opendataintegration,koutras2021valentine}, most harmonization processes still require significant manual effort, domain expertise, and iterative refinement. These challenges become even more pronounced in biomedical integration scenarios, where datasets differ substantially in structure and semantics. For example, in a recent proteogenomic analysis, \citet{li2023proteogenomic} aggregated data from ten diverse cancer studies to map them onto the National Cancer Institute's standardized Genomic Data Commons (GDC) model\footnote{\url{https://portal.gdc.cancer.gov/}}. This integration proved challenging due to heterogeneity across the 700+ attributes of the target model, requiring both schema alignment and diverse transformation operations, ranging from textual modifications to complex numerical conversions. Currently, no single automated solution is capable of resolving this wide spectrum of heterogeneity~\cite{magneto@vldb2025}.

BDI-Kit is an open-source system\footnote{\url{https://github.com/VIDA-NYU/bdi-kit}\label{fn:repo}} designed to support data harmonization as an interactive, human-in-the-loop process. Rather than aiming for fully automated integration, BDI-Kit provides a diverse set of matching primitives that generate candidate matches, allowing users to inspect and refine results. The system explicitly treats harmonization as an exploratory workflow in which automated methods and human judgment are tightly coupled.

Our prior work \cite{lopez2026bdikit} presents the design and algorithms of BDI-Kit as a general-purpose toolkit for data harmonization. In contrast, this paper focuses on demonstrating how users interact with the system in practice. The goal of the demonstration is to show how harmonization workflows progress step by step, how users reason about intermediate results, and how human feedback influences the final outcome.

To this end, we present BDI-Kit through two complementary interaction scenarios \revision{(see the demo video\footref{fn:repo})}. The first one shows how data scientists can use the Python API to harmonize two tables by composing schema and value matching primitives and iteratively refining the results. The second scenario highlights how domain experts can perform table-to-model harmonization using a conversational interface, where an AI agent orchestrates primitives in response to natural-language requests. 

%% file: sections/02_system.tex
\section{BDI-Kit Overview}
BDI-Kit is an open-source interactive library for data harmonization that can be installed via PyPI (\texttt{pip install bdi-kit}). It integrates schema and value matching techniques with explicit support for human-in-the-loop refinement. The system is designed around the observation that harmonization is rarely a fully automated process: users must inspect intermediate results, resolve ambiguities, and iteratively refine matches before producing an integrated dataset.

\begin{figure}[t]
    \centering
    \includegraphics[width=0.48\textwidth]{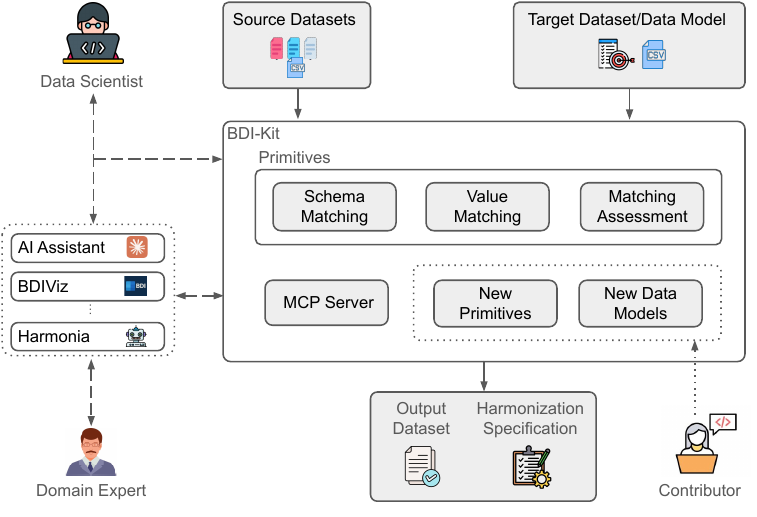}
    \caption{Users harmonize source data to target tables or data models via schema and value matching primitives, through a Python API or AI-assisted interfaces, producing a harmonized dataset and reusable specification.}
    \label{fig:library_workflow}
\end{figure}

BDI-Kit exposes a set of composable harmonization primitives. These primitives operate on tabular data and can be orchestrated programmatically or invoked indirectly through an AI-assisted conversational interface. The system is extensible, allowing contributors to integrate new matching algorithms, additional data models, and target schemas. The output of the harmonization process is a harmonized dataset with a harmonization specification that makes transformations explicit and supports reuse across datasets. Figure \ref{fig:library_workflow} illustrates the system architecture and interaction flow.

\subsection{Data Model and Inputs}
BDI-Kit operates on datasets represented as pandas DataFrames. A harmonization task is defined by: (i) a source dataset, provided as a DataFrame; (ii) a target, which can be either another DataFrame or a predefined data model.

Target data models are represented internally through a lightweight abstraction that exposes attribute names, permissible values, and metadata. This abstraction allows BDI-Kit to support heterogeneous standards without imposing a fixed schema representation. \revision{In this work, data model refers to domain-specific schemas (e.g., GDC), not abstract database models such as relational or graph}.

\subsection{Harmonization Primitives}
The system provides three categories of primitives that form the building blocks of harmonization workflows.

\para{Schema Matching} BDI-Kit integrates 12 schema matching primitives that identify candidate matches between attributes of the source  and target datasets, including traditional methods, algorithmic solutions, and LLM-based approaches. Each invocation produces candidate matches with similarity scores, enabling users to inspect alternatives and reason about ambiguous cases.

\para{Value Matching} Value matching primitives operate on pairs of matched attributes and identify equivalent values. BDI-Kit supports 5 value matching strategies, including textual similarity, embedding-based similarity, and numeric transformations. The output is a set of candidate value matches that can be inspected and refined.

\revision{\para{Matching Assessment} These primitives evaluate and explain matches generated by the algorithms. These explanations clarify why a match was made and whether it is valid or questionable. This enables users to identify errors and improve the matching process.}

All primitives are designed to be composable: the output of one primitive can be passed as input to another, enabling flexible and incremental workflows. \revision{Users can compose primitives freely and apply them to any attribute or value; BDI-Kit does not automatically select functions by type, leaving these decisions to the user. Also, BDI-Kit handles moderate-size datasets efficiently, though very large datasets may require optimized or parallelized primitives.}

\subsection{Harmonization Specification}
BDI-Kit represents the outcome of a harmonization process through a harmonization specification, a declarative artifact that explicitly defines how a source dataset is transformed into a target schema or data model. \revision{Currently, BDI-Kit focuses on one-to-one attribute correspondences rather than full schema mappings (e.g., GaV or LaV).} The specification captures the final set of attribute correspondences, the value-level transformations applied to align representations, and any user-provided refinements that override or complement automated matches.

The harmonization specification also serves as a reusable representation of integration knowledge. Once created, it can be applied to other datasets that share the same or a similar source schema, eliminating the need to repeat matching and refinement steps. This is particularly useful in recurring harmonization tasks, where the same transformations must be applied across multiple datasets.

\subsection{Extensibility and Interoperability}
BDI-Kit is designed to be extensible at both the data model and algorithmic levels. New target data models can be incorporated by providing lightweight schema definitions and metadata, enabling the system to support additional standards and domain-specific representations without modifying core components. This design allows BDI-Kit to evolve alongside emerging data models and integration requirements. The system also supports extensibility through pluggable harmonization primitives. Developers can add new schema matching or value matching methods, enabling experimentation with alternative methods.

BDI-Kit also exposes its functionality through the Model Context Protocol (MCP), allowing external AI agents to interact with the system in a model-agnostic manner. Through MCP, BDI-Kit can be orchestrated by different AI assistants, enabling harmonization workflows to be embedded into broader AI-driven data management pipelines while preserving user control and system transparency.

\begin{figure*}[h!]
    \centering
    \includegraphics[width=0.95\textwidth]{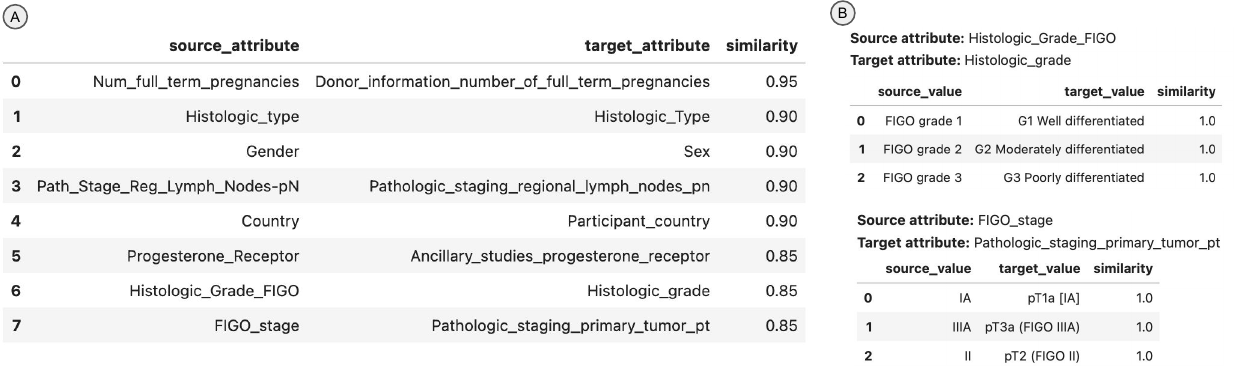}
    \caption{Results of calling \texttt{match\_schema()} and \texttt{match\_values()} functions via the Python API.}
    \label{fig:python_api_results}
\end{figure*}

\subsection{Interaction Modalities}
BDI-Kit supports two complementary interaction modalities that expose the same underlying system functionality.

\para{Python API} The Python API enables data scientists to embed harmonization workflows into data processing pipelines. Users can programmatically invoke primitives, inspect intermediate results, \revision{review proposed match pairs before applying transformations,} and introduce custom refinements. This interface emphasizes reproducibility and integration with existing analytical workflows.

\para{AI-Assisted Conversational Interface} For users with limited programming experience, BDI-Kit can be accessed through an AI-assisted interface built on MCP. In this mode, an AI agent interprets natural-language requests, selects appropriate primitives, and executes them on behalf of the user. Importantly, the agent does not replace BDI-Kit’s logic; instead, it acts as an orchestration and explanation layer. \revision{To ensure safe and reliable harmonization, the system enforces guardrails: all automated suggestions report similarity scores, provenance flows show decision rationale, and users retain final control to accept, modify, or reject matches. Users can also challenge, add constraints, or revise suggestions at any step.}

%% file: sections/03_demonstration.tex
\section{Demonstration Scenarios}

\subsection{Scenario 1: Python API Harmonization}
\revision{This scenario demonstrates how BDI-Kit supports interactive and reproducible table-to-table harmonization {between two endometrial tumor-related datasets (from \citep{dou2020proteogenomic} to \citep{dou2023proteogenomic})} through its Python API. The workflow enables researchers to iteratively perform schema and value matching with human-in-the-loop refinement, while producing reusable harmonization specifications.}

\revision{\para{Dataset Preparation} To demonstrate both harmonization and reuse, we partition the source dataset into two subsets: (i) a base subset, used to construct the harmonization specification, and (ii) a held-out subset, used to simulate a new incoming dataset. The target remains unchanged.}

\para{Schema Matching} \revision{The researcher invokes \texttt{\seqsplit{match\_schema()}}, whose outputs are shown in Figure~\ref{fig:python_api_results}A.} The results illustrate that BDI-Kit captures a variety of semantic relationships, including synonym matches such as \textit{Gender} → \textit{Sex}, and closely related clinical concepts such as \textit{Histologic\_Grade\_FIGO} → \textit{Histologic\_grade}. The ranked correspondences and similarity scores allow the researcher to quickly assess match quality and identify potential ambiguities. Since the proposed schema matches are largely correct, the workflow proceeds to value-level alignment.

\para{Value Matching} After confirming the schema matches, the researcher invokes the \texttt{match\_values()} function to align the values of each matched attribute pair. Figure~\ref{fig:python_api_results}B shows some results of this step. For example, within the match \textit{Histologic\_Grade\_FIGO} → \textit{Histologic\_grade}, the source value \textit{FIGO grade 2} is correctly matched to the target value \textit{G2 Moderately differentiated}. 

\para{Human-in-the-Loop Correction} The demonstration also highlights the role of expert validation in data harmonization. For the attribute pair \textit{FIGO\_stage} → \textit{Pathologic\_staging\_primary\_tumor\_pt}, the automatically proposed  value match \textit{IA}  →  \textit{pT1a[IA]} is identified as incorrect. Using the editable interface, the researcher corrects this match by leveraging consistency across the other correctly matched values, updating the target value: \textit{IA}  →  \textit{pT1a (FIGO IA)}. This interaction illustrates how BDI-Kit combines automated matching with efficient manual refinement when ambiguities arise.

\begin{figure*}[h!]
    \centering
    \includegraphics[width=0.95\textwidth]{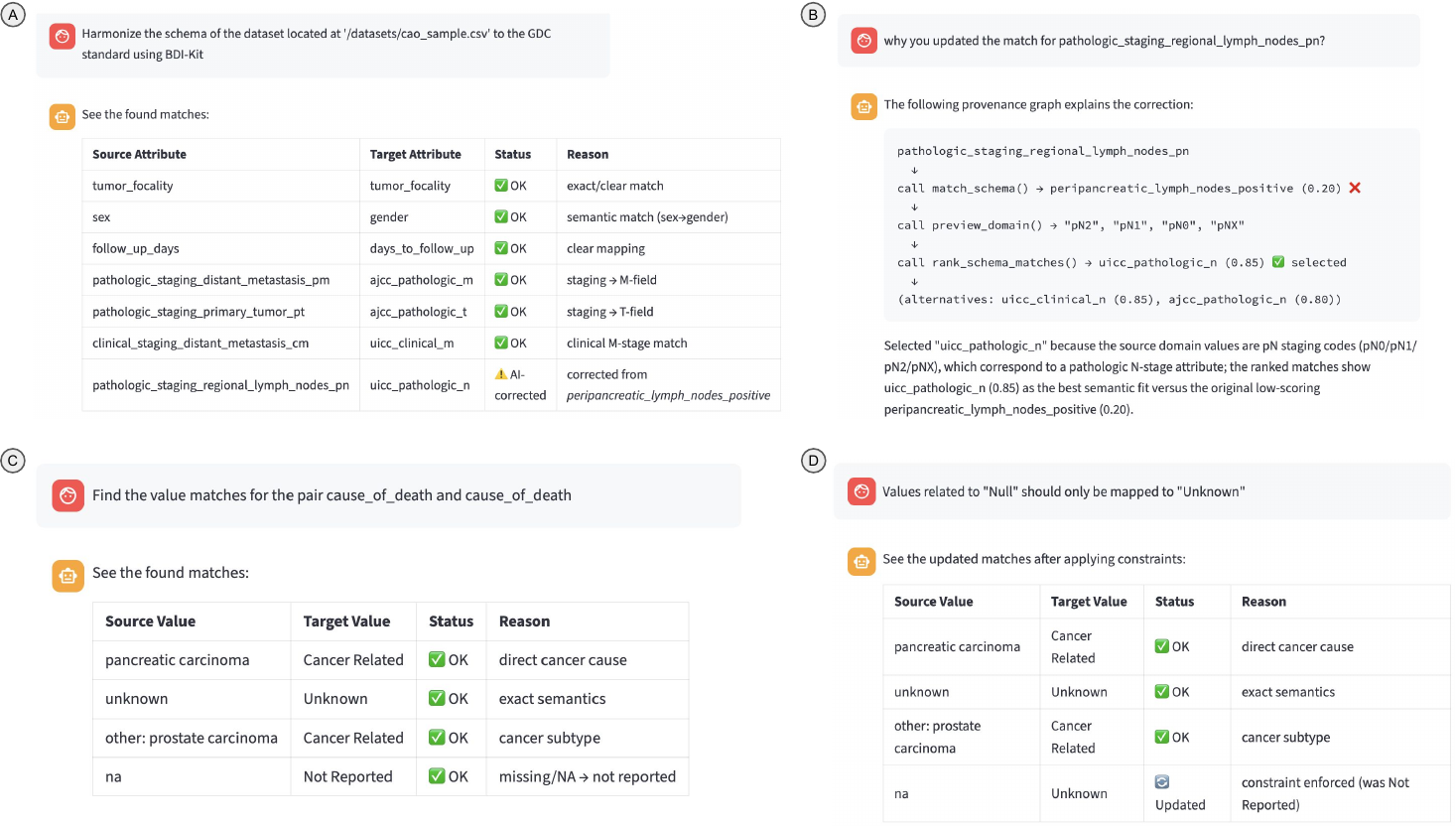}
    \caption{BDI-Kit accessed through an AI agent orchestrating schema and value matching.}
    \label{fig:chat_interface_results}
\end{figure*}

\revision{\para{Harmonization Specification Generation and Reuse} The system produces a harmonization specification} that captures both schema-level and value-level correspondences in a declarative JSON format. A snippet of this specification is shown below. Each entry explicitly defines how source attributes and their values are transformed to conform to the target schema. \revision{As a final step, the specification is applied to the held-out subset. Without recomputing matches, BDI-Kit reuses the learned mappings to transform the data into the target representation.}

\definecolor{punct}{RGB}{153, 0, 0}
\definecolor{delim}{RGB}{20,105,176}
\definecolor{my_lightgray}{RGB}{242,242,242}
\lstdefinelanguage{json}{
    basicstyle=\ttfamily\scriptsize,
    numbers=none,
    showstringspaces=false,
    breaklines=true,
    frame=none,
    backgroundcolor=\color{my_lightgray},
    literate=
     *{0}{{{\color{black}0}}}{1}
      {1}{{{\color{black}1}}}{1}
      {2}{{{\color{black}2}}}{1}
      {3}{{{\color{black}3}}}{1}
      {4}{{{\color{black}4}}}{1}
      {5}{{{\color{black}5}}}{1}
      {6}{{{\color{black}6}}}{1}
      {7}{{{\color{black}7}}}{1}
      {8}{{{\color{black}8}}}{1}
      {9}{{{\color{black}9}}}{1}
      {:}{{{\color{punct}{:}}}}{1}
      {,}{{{\color{punct}{,}}}}{1}
      {\{}{{{\color{delim}{\{}}}}{1}
      {\}}{{{\color{delim}{\}}}}}{1}
      {[}{{{\color{delim}{[}}}}{1}
      {]}{{{\color{delim}{]}}}}{1},
    stringstyle=\color{black},  
    commentstyle=\color{black}, 
    keywordstyle=\color{black}, 
}

\begin{lstlisting}[language=json]
   {"source_attribute": "Histologic_Grade_FIGO",
    "target_attribute": "Histologic_grade",
    "mapper": {
      "FIGO grade 1": "G1 Well differentiated",
      "FIGO grade 2": "G2 Moderately differentiated",
      "FIGO grade 3": "G3 Poorly differentiated"}},
   {"source_attribute": "FIGO_stage",
    "target_attribute": "Pathologic_staging_primary_tumor_pt",
    "mapper": {
      "IA": "pT1a (FIGO IA)",
      "IIIA": "pT3a (FIGO IIIA)",
      "II": "pT2 (FIGO II)"}}
\end{lstlisting}

\subsection{Scenario 2: AI-Assisted Harmonization}
This demonstration presents a conversational, AI-assisted interface for harmonizing a \revision{tabular pancreatic cancer dataset against a data model (from \citep{cao2021proteogenomic} to \textit{GDC})}. Through natural-language interaction, the AI agent guides the user across schema matching, validation, and value matching, combining automated recommendations with human review to support efficient and transparent harmonization.

\revision{
\para{AI-Assisted Match Review}
After invoking schema harmonization, the conversational agent presents the discovered matches in a structured table, clearly indicating whether each match was accepted as-is or automatically corrected (see Figure \ref{fig:chat_interface_results}A). Matches identified as reliable,  such as  \textit{tumor\_focality} or \textit{sex} are marked as \textit{OK}, while those adjusted by the agent are labeled \textit{AI-corrected} (e.g. \textit{pathologic\_staging\_regional\_lymph\_nodes\_pn}), along with a concise reason. This allowed the user to quickly review the harmonization results and understand where the agent intervened. The interface then provides actionable next steps, such as accepting the matches or proceeding with value matching.

\para{Provenance-Aware Explanations}
When the user requests clarification for a corrected match, the agent provides a compact provenance flow that summarizes the decision process. The explanation is presented as a vertical flow showing the initial match, domain inspection (via \texttt{preview\_domain()}), alternative ranking (via \texttt{rank\_schema\_matches()}), and the final selected attribute. This provenance graph is followed by a brief explanation of why the final match was chosen, helping the user understand the AI’s reasoning behind the corrections (Figure \ref{fig:chat_interface_results}B).

\para{Interactive Constraint-Based Matching}
To illustrate interactive \textit{what-if} scenarios, the user performed value matching for the attribute pair \textit{cause\_of\_death} → \textit{cause\_of\_death}. The AI agent first generated initial matches with short semantic justifications (Figure \ref{fig:chat_interface_results}C). The user then introduced a constraint specifying that values related to null should only map to \textit{Unknown}. The agent then identified the affected entries, and updated the results accordingly, changing \textit{na} from \textit{Not Reported} to \textit{Unknown}, while leaving the remaining matches unchanged (Figure \ref{fig:chat_interface_results}D). This experiment demonstrates how user-defined constraints can dynamically refine harmonization results and make the integration process more interactive and transparent.
}

%% file: sections/04_acknowledgments.tex
\revision{\section*{Acknowledgments}
This work was supported in part by DARPA ASKEM (HR0011262087), ARPA-H BDF, and NSF (OAC-2411221). The views, opinions, and findings expressed are those of the authors and should not be interpreted as representing the views or policies of these agencies.}

%% file: references.bib
@article{magneto@vldb2025,
  author = {Liu, Yurong and Pena, Eduardo H. M. and Santos, A\'{e}cio and Wu, Eden and Freire, Juliana},
  title = {Magneto: Combining Small and Large Language Models for Schema Matching},
  year = {2025},
  publisher = {VLDB Endowment},
  volume = {18},
  number = {8},
  issn = {2150-8097},
  journal = {Proceedings of the VLDB Endowment (PVLDB)},
  pages = {2681--2694},
  numpages = {14}
}

@inproceedings{koutras2021valentine,
  title={{Valentine: Evaluating Matching Techniques for Dataset Discovery}},
  author={Koutras, Christos and Siachamis, George and Ionescu, Andra and Psarakis, Kyriakos and Brons, Jerry and Fragkoulis, Marios and Lofi, Christoph and Bonifati, Angela and Katsifodimos, Asterios},
  booktitle={Proceedings of International Conference on Data Engineering (ICDE)},
  pages={468--479},
  year={2021}
}

@article{dou2020proteogenomic,
  title={{Proteogenomic Characterization of Endometrial Carcinoma}},
  author={Dou, Yongchao and Kawaler, Emily and Zhou, Daniel Cui and Gritsenko, Marina and Huang, Chen and Blumenberg, Lili and Karpova, Alla and Petyuk, Vladislav and others},
  journal={Cell},
  volume={180},
  number={4},
  pages={729--748},
  year={2020},
  publisher={Elsevier}
}

@article{dou2023proteogenomic,
  title={{Proteogenomic Insights Suggest Druggable Pathways in Endometrial Carcinoma}},
  author={Dou, Yongchao and Katsnelson, Lizabeth and Gritsenko, Marina and Hu, Yingwei and Reva, Boris and Hong, Runyu and Wang, Yi-Ting and others},
  journal={Cancer Cell},
  volume={41},
  number={9},
  pages={1586--1605},
  year={2023},
  publisher={Elsevier}
}

@book{doan2012integrationBook,
author = {Doan, AnHai and Halevy, Alon and Ives, Zachary},
title = {Principles of Data Integration},
year = {2012},
isbn = {0124160441},
publisher = {Morgan Kaufmann Publishers Inc.},
edition = {1st},
}

@article{cafarellaIntegrationonweb2009,
author = {Cafarella, Michael J. and Halevy, Alon and Khoussainova, Nodira},
title = {{Data Integration for the Relational Web}},
year = {2009},
volume = {2},
number = {1},
issn = {2150-8097},
journal = {Proceedings of the VLDB Endowment (PVLDB)},
OPTjournal = {Proceedings of the International Conference on Very Large Data Bases (PVLDB)},
OPTmonth = aug,
pages = {1090–1101},
numpages = {12}
}

@article{miller2018opendataintegration,
author = {Miller, Ren\'{e}e},
title = {{Open Data Integration}},
year = {2018},
volume = {11},
number = {12},
issn = {2150-8097},
OPTbooktitle = {Proceedings of the International Conference on Very Large Data Bases (PVLDB)},
journal = {Proceedings of the VLDB Endowment (PVLDB)},
pages = {2130–2139},
numpages = {10}
}

@article{lopez2026bdikit,
  title={{BDI-Kit: An AI-Powered Toolkit for Biomedical Data Harmonization}},
  author={Lopez, Roque and Santos, A{\'e}cio and Koutras, Christos  and Freire, Juliana},
  journal={{Patterns}},
  volume={7},
  pages={},
  year={2026},
  publisher={}
}

@article{li2023proteogenomic,
  title={{Proteogenomic Data and Resources for Pan-cancer Analysis}},
  author={Li, Yize and Dou, Yongchao and Leprevost, Felipe Da Veiga and Geffen, Yifat and Calinawan, Anna and Aguet, Fran{\c{c}}ois and Akiyama, Yo and others},
  journal={Cancer Cell},
  volume={41},
  number={8},
  pages={1397--1406},
  year={2023},
  publisher={Elsevier}
}

@article{cao2021proteogenomic,
title = {{Proteogenomic Characterization of Pancreatic Ductal Adenocarcinoma}},
journal = {Cell},
volume = {184},
number = {19},
pages = {5031-5052},
year = {2021},
issn = {0092-8674},
  author={Cao, Liwei and Huang, Chen and Zhou, Daniel Cui and Hu, Yingwei and Lih, Mamie and Savage, Sara and Krug, Karsten and Clark, David and others},
}
